\documentclass{article}
\usepackage{spconf,amsmath,graphicx, booktabs, multirow, tabularx,makecell}
\usepackage{amssymb}
\usepackage[
    left=0.753in,
    right=0.727in,
    top=1.007in,
    bottom=1.04in
]{geometry}

\usepackage{enumitem}
\setlist{nosep, leftmargin=14pt}

\usepackage{mwe} 

\title{CCIS-Diff: A Generative Model with Stable Diffusion Prior for Controlled Colonoscopy Image Synthesis}
\name{Yifan Xie$^{1,4}$, Jingge Wang$^{2,3}$, Tao Feng$^{2}$, Fei Ma$^{4*}$, Yang Li$^{2,3*}$\thanks{$^{*}$Corresponding email: mafei@gml.ac.cn, yangli@sz.tsinghua.edu.cn}}
\address{$^1$School of Software Engineering, Xi'an Jiaotong University, Xi'an, China \\$^2$Tsinghua Shenzhen International Graduate School, Tsinghua University, Shenzhen, China \\$^3$Shenzhen Key Laboratory of Ubiquitous Data Enabling, Shenzhen, China
\\$^4$Guangdong Laboratory of Artificial Intelligence and Digital Economy (SZ), Shenzhen, China}
\begin{document}
%
\maketitle
\begin{abstract}
Colonoscopy is crucial for identifying adenomatous polyps and preventing colorectal cancer. However, developing robust models for polyp detection is challenging by the limited size and accessibility of existing colonoscopy datasets. 
While previous efforts have attempted to synthesize colonoscopy images, current methods suffer from instability and insufficient data diversity. Moreover, these approaches lack precise control over the generation process, resulting in images that fail to meet clinical quality standards.
To address these challenges, we propose $\textbf{CCIS-DIFF}$, a $\textbf{C}$ontrolled generative model for high-quality $\textbf{C}$olonoscopy $\textbf{I}$mage $\textbf{S}$ynthesis based on a $\textbf{Diff}$usion architecture.
Our method offers precise control over both the spatial attributes (polyp location and shape) and clinical characteristics of polyps that align with clinical descriptions. Specifically, we introduce a blur mask weighting strategy to seamlessly blend synthesized polyps with the colonic mucosa, and a text-aware attention mechanism to guide the generated images to reflect clinical characteristics.
Notably, to achieve this, we construct a new multi-modal colonoscopy dataset that integrates images, mask annotations, and corresponding clinical text descriptions. 
Experimental results demonstrate that our method generates high-quality, diverse colonoscopy images with fine control over both spatial constraints and clinical consistency, offering valuable support for downstream segmentation and diagnostic tasks.

\end{abstract}
\section{Introduction}
\label{sec:Introduction}
Colonoscopy is an essential tool for detecting adenomatous polyps and reducing rectal cancer mortality rates \cite{fan2020pranet}. 
However, 
training models for automatic polyp detection is challenging due to the small scale of available colonoscopy datasets, making it difficult to have sufficient robustness and generalization that meet real-world clinical demands.

\begin{figure}[t]
    \centering
    \includegraphics[width=0.85\linewidth]{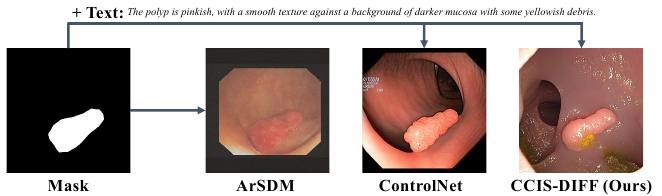}
    \vskip -10pt
    \caption{
    In contrast to ArSDM \cite{du2023arsdm} and ControlNet \cite{zhang2023adding}, CCIS-DIFF can utilize not only mask but also textual description to generate high-fidelity, text-consistent colonoscopy images.
    }
    \label{fig:figure1}
    \vskip -10pt
\end{figure}

To address this problem, previous methods \cite{ma2020cycle,xu2020ofgan,du2023arsdm,jeong2024uncertainty} primarily relied on generative adversarial networks or diffusion models to synthesize more colonoscopy images. 
Although these efforts aim to address the data scarcity problem, they struggle to generate a sufficiently diverse and high-quality image, and the generation process lacks adequate control, leading to images that fail to meet clinical requirements for practical use.
As illustrated in Fig. \ref{fig:figure1}, ArSDM \cite{du2023arsdm} only utilizes the mask to synthesize the colonoscopy image and the generated image is of poor quality and contains noise.
Meanwhile, large-scale text-to-image (T2I) diffusion models such as Stable Diffusion \cite{rombach2022high} and DALL·E \cite{ramesh2022hierarchical} have demonstrated remarkable capabilities in generating images from various prompts. This raises an important question: Can colonoscopy images be generated in a controlled manner using a pre-trained large-scale T2I model? 
In response, we present an innovative generative method to synthesize high-quality colonoscopy images in a controlled manner.
\begin{figure*}[t]
\centering
\includegraphics[width=0.82\textwidth]{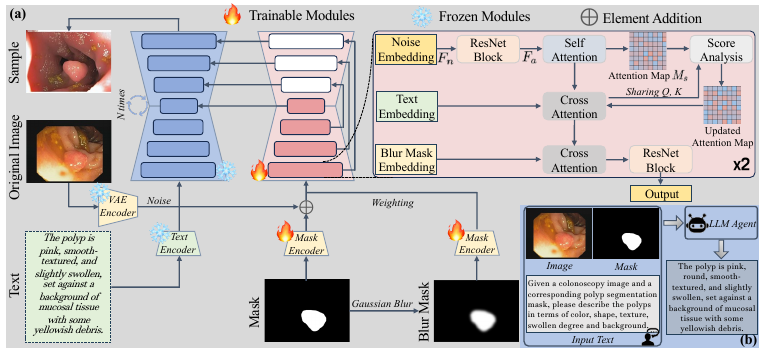} 
\vskip -10pt
\caption{(a) The overall architecture of CCIS-DIFF. 
The parameters of the frozen modules refer to Stable Diffusion v1.5 \cite{rombach2022high}. 
The trainable modules incorporate a text-aware attention mechanism to address the issue of neglecting text prompts. Additionally, we introduce a blur mask weighting strategy to ensure seamless integration of synthesized polyps with the colonic mucosa.
(b) The construction pipeline of the multi-modal colonoscopy dataset.}
\label{fig:figure2}
\vskip -10pt
\end{figure*}
The main contributions of this paper can be summarized as follows:
\begin{itemize}
\item We propose a novel generative model, named CCIS-DIFF, 
which offers fine control over both the spatial attributes and clinical characteristics of polyps, enabling more clinically consistent image synthesis for practical use.
\item We introduce the blur mask weighting strategy to ensure seamless integration of synthesized polyps with the colonic mucosa, along with a text-aware attention mechanism that incorporates textual information into the generation process, to enable customization of polyp images.
\item We design the first multi-modal colonoscopy dataset that uniquely combines colonoscopy images, segmentation masks, and clinical text descriptions, providing precise alignment between visual data and clinical information for improved training and evaluation.
\end{itemize}

\section{Method}
\label{sec:Method}


\subsection{Multi-Modal Colonoscopy Dataset}
A key challenge for current diffusion models in colonoscopy image synthesis \cite{du2023arsdm,jeong2024uncertainty} is the absence of 
a dataset with consistent spatial characteristics of polyps and their corresponding clinical textual descriptions. Existing datasets, such as EndoScene \cite{vazquez2017benchmark}, CVC-ClinicDB \cite{bernal2015wm}, and Kvasir \cite{jha2020kvasir}, primarily contain image-mask pairs that focus on polyp regions. However, these datasets do not include detailed clinical text descriptions, limiting their usefulness for training models that aim to generate clinically accurate and diverse synthetic images.
To address this limitation, we built a new multi-modal colonoscopy dataset based on existing datasets, containing three key components: colonoscopy images, segmentation masks, and clinical text descriptions. 
Such a dataset is vital for fine-tuning pretrained diffusion models, allowing them to adapt effectively to controlled colonoscopy image synthesis.

An overview of our dataset construction process is shown in Fig. \ref{fig:figure2} (b).
To generate accurate captions that accurately reflect both spatial constraints and relevant clinical characteristics, we construct the open-source LLaMA \cite{touvron2023llama} large language model (LLM) agent. This agent takes a textual prompt and an image-mask colonoscopy pair to create captions for both the foreground and background. To increase textual diversity, the LLM incorporates different aspects like color, shape, texture, and swelling. As a result, we develop a multi-modal colonoscopy dataset consisting of triplets of colonoscopy images, mask images, and their corresponding textual descriptions which provides the necessary foundation for generating images with improved realism and variability.

\subsection{CCIS-DIFF Architecture}
\subsubsection{Overview}
The overview architecture of our CCIS-DIFF is presented in Fig. \ref{fig:figure2} (a).
Using our multi-modal colonoscopy dataset, we provide the original colonoscopy image $I$, the mask image $M$, and the corresponding text description $T$. Each of these components passes through its respective encoder, noting that the text encoder is frozen. 
Additionally, we develop a blur mask to ensure that the generated polyp integrates seamlessly with the background, the detailed description will be illustrated in Sec. \ref{blur}.

Based on ControlNet \cite{zhang2023adding}, we adopt the trainable diffusion branch and implement the zero convolution strategy to protect this branch by eliminating random noise as gradients in the initial training steps. 
This structure, when applied to large models like Stable Diffusion \cite{rombach2022high}, enables the frozen parameters to preserve the integrity of the production-ready model that has been trained on billions of images. Meanwhile, the trainable diffusion branch leverages this large-scale pre-trained model to establish a robust backbone for managing multi-modal input conditions. 
Furthermore, to address the issue of neglecting text prompt and effectively incorporating textual information into the generation process, we incorporate a text-aware attention mechanism (Sec. \ref{text}) in the trainable diffusion branch.

\subsubsection{Blur Mask Weighting Strategy}
\label{blur}
The purpose of the blur mask weighting strategy is to ensure that the generated polyp is seamlessly integrated with the background. 
To achieve this, we apply a Gaussian blur operation $\sigma$ to the mask image $M$, softening the transition between the polyp mask region and the background. 
We utilize two separate mask encoders to extract features from the mask and the blurred mask, each with non-shared parameters. 
Subsequently, a weighting matrix $M_w$ is introduced to balance the two mask branches, and $M_w$ is learned using a three-layer MLP.
Thus the blur mask embedding $F_b$ can be constructed as:
\begin{equation}
F_b = M_w \odot E_m (\sigma(M)),
\end{equation}
where $\odot$ is the Hadamard product and $E_m$ denotes the mask encoder. 

\subsubsection{Text-Aware Attention Mechanism}
\label{text}
In our experiments, we observed that existing methods, such as ControlNet \cite{zhang2023adding}, often overlook the text prompt and rely more heavily on the mask image. We hypothesize that this visual dominance over the text prompt arises from the text-free nature of the self-attention layers. To address this issue, we propose a text-aware attention mechanism that leverages the cross-attention matrix to regulate the output of the self-attention.

Given the noise embedding $F_n$, it is first passed through a ResNet block, as referenced in ControlNet \cite{zhang2023adding}. Following this, we obtain the attention input tensor $F_a \in \mathbb{R}^{(h \times w) \times d}$, which is processed through projection layers to derive the queries $Q_s$, keys $K_s$, and values $V_s$, as well as the attention map $M_{s} = \frac{Q_s K_s^{T}}{\sqrt{d}} \in \mathbb{R}^{hw\times hw}$. To reduce the strong influence of the mask image, we adjust the attention scores of the prompt text. Specifically, we begin by constructing the cross-attention similarity matrix:
\begin{equation}
M_{c}=SoftMax(Q_{c}K_{c}^{T}/\sqrt{d}),
\end{equation}
where $Q_{c} \in \mathbb{R}^{(h\times w)\times d}, K_{c} \in \mathbb{R}^{l\times d}$ represent the query and key tensors from their corresponding cross-attention layers,
and $l$ denotes the number of tokens in the text prompt.
For each pixel $j$, we define its similarity to the text prompt by summing its similarity scores with the embedding indexed by $M_c$. We then apply a clamp operation to normalize the scores $s_j$ within the range $\left[0,1\right]$:
\begin{equation}
s_j = Norm (Sum_{j}(M_c)).
\end{equation}

By calculating the scores for each pixel, we can obtain the final text-aware map $S$. 
Subsequently, we utilize $S$ to compute the updated attention map and refine the self-attention process.

Finally, the output of the cross-attention mechanism interacts once more with the blur mask embedding, generating the final output embedding through a ResNet block.

\subsubsection{Training and Inference}

During the training phase, the diffusion models start with an original image $I$ and progressively add noise to create a noisy embedding $I_t$, where $t$ indicates the number of times noise is added. 
The models learn a network $\epsilon_{\theta}$ to predict the noise added to the noisy image $I_t$, based on a set of conditions that include the time step $t$, a text prompt $T$, and a mask image $M$.
The overall loss function $\mathcal{L}$ for the entire diffusion model is represented as:
\begin{equation}
\mathcal{L}=\mathbb{E}_{I,t,T,M,\epsilon\sim\mathcal{N}(0,1)}\Big[\|\epsilon-\epsilon_{\theta}(I_{t},{t},T,M))\|_{2}^{2}\Big].
\end{equation}
Additionally, we randomly replace half of the text prompts with empty strings. This strategy enhances the model's ability to directly recognize the semantics of the input mask image, serving as a substitute for the text prompt.

During the inference phase, the diffusion models automatically generate noise and produce the final colonoscopy images based on a text prompt and a mask image. Simultaneously, we utilize Classifier-Free Guidance (CFG) \cite{ho2022classifier} to enhance the sampling process.

\begin{table}[t]
    \caption{The quantitative results of the colonoscopy image synthesis. The boldface indicates the best performance.}
    \resizebox{\linewidth}{!}{%
    \begin{tabular}{lccc}
        \toprule
         Method      & FID $\downarrow$ & CLIP-score $\uparrow$ & CLIP-image $\uparrow$ \\
        \midrule
       ControlNet  &87.32 &30.78  &87.09    \\
Uni-ControlNet &92.74 &30.57  &86.46 \\
\hline
Ours  & \textbf{71.73} & \textbf{31.96} & \textbf{88.70}\\ 
         \bottomrule
    \end{tabular}%
    }
    \vspace{-0.37cm}
	\label{table1}
\end{table}

\begin{figure*}[t]
\centering
\includegraphics[width=0.85\textwidth]{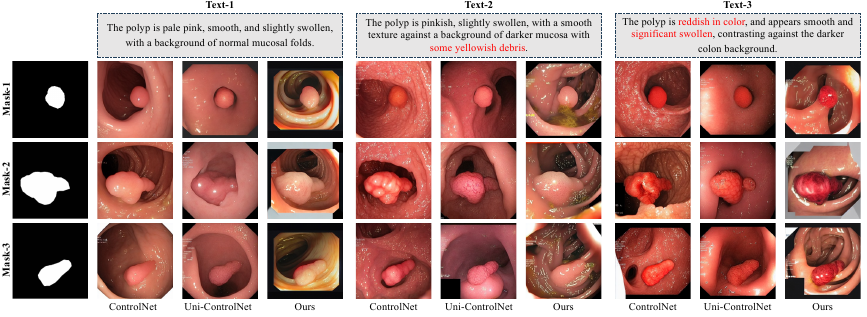} 
\vskip -15pt
\caption{Qualitative comparison of colonoscopy image synthesis.}
\label{fig:figure3}
\vskip -15pt
\end{figure*}

\section{Experiment and results}
\label{sec:Experiment}


\subsection{Implementation Details}
All methods were implemented in PyTorch on a single NVIDIA A100 40G GPU. We utilized the pre-trained Stable Diffusion v1.5 \cite{rombach2022high} model, following ControlNet \cite{zhang2023adding}, to replicate its UNet encoder as the trainable copy encoder. The batch size was configured to 4, and the learning rate was set to 1e-5. For inference, we used a default CFG scale of 7.0. The DDIM sampler was employed, using 20 steps to sample each image.

\subsection{Baseline Algorithms and Evaluation Metrics}
For the colonoscopy image synthesis task, we compared our method with ControlNet \cite{zhang2023adding} and Uni-ControlNet \cite{zhao2024uni}, all of which were trained on our multi-modal colonoscopy dataset. 
Following \cite{zhang2023adding}, we assessed performance using FID, CLIP-score, and CLIP-image (measuring the similarity between the generated image and the reference image).

For the downstream polyp segmentation task, 
we utilized PraNet \cite{fan2020pranet} and Polyp-PVT \cite{dong2021polyp} as baseline segmentation models with their default settings, and assessed their performance using mean Dice (mDice) and mean Intersection over Union (mIoU) metrics.



\begin{table}[t]
    \caption{User study. The rating scale ranges from 1 to 5, with higher numbers indicating better performance.}
    \resizebox{\linewidth}{!}{%
    \begin{tabular}{lccc}
        \toprule
         Method      & Image Fidelity & Mask Accuracy & Text Accuracy \\
        \midrule
       ControlNet  &3.852 &3.962  &3.918    \\
Uni-ControlNet &2.904 &2.966  &3.392 \\
\hline
Ours  & \textbf{4.266} & \textbf{4.466} & \textbf{4.518}\\
         \bottomrule
    \end{tabular}%
    }
    \vspace{-0.37cm}
	\label{table2}
\end{table}

\begin{table}[]
\caption{Comparisons of different settings applied on two polyp segmentation baselines across three public datasets.}
\resizebox{\linewidth}{!}{
\begin{tabular}{@{}lcccccc@{}}
\toprule
 & \multicolumn{2}{c}{EndoScene}       & \multicolumn{2}{c}{CVC-ClinicDB}     & \multicolumn{2}{c}{Kvasir}  \\ \cmidrule(l){2-7} 
Settings               & mDice         & mIoU        & mDice          & mIoU      & mDice         & mIoU  \\ 
\midrule

PraNet         & 86.1         & 79.1         & 90.8         & 86.1    & 89.2  & 84.0    \\
PraNet + ControlNet      & 85.4           & 77.7       & 90.1         & 85.1    & 90.6  & 85.2     \\
PraNet + Uni-ControlNet      & 85.9    & 78.2       & 89.5        &85.1    & 87.9  & 82.5     \\
PraNet + Ours      & \textbf{88.5}           & \textbf{81.0}        & \textbf{92.8}          & \textbf{88.0}     & \textbf{92.8}   & \textbf{87.1}      \\
\midrule
PVT    & 88.2    & 81.2   & 92.8         & 87.6  & 91.5  & 86.8       \\
PVT + ControlNet      & 88.3          &81.4         & 92.7  & 88.2  & 90.3  & 84.8        \\
PVT + Uni-ControlNet       & 85.8          &79.6        & 89.3  & 83.8 & 86.5  & 80.7         \\
PVT + Ours    & \textbf{89.3}         &\textbf{82.1}         & \textbf{93.5}  & \textbf{88.4}  & \textbf{92.4}  & \textbf{87.0}         \\
\bottomrule
\end{tabular}
}
\setlength{\abovecaptionskip}{0.25cm}
\label{table3}
\vspace{-0.5cm}
\end{table}

\subsection{Results}
The quantitative results of the colonoscopy image synthesis are illustrated in Table \ref{table1}. 
Our method outperforms others in both visual quality and text alignment.
Furthermore, we present a comparison of our method with others in Fig. \ref{fig:figure3}. The results show that the polyp region generated by our method is more consistent with the mask image, and the overall information aligns more closely with the text prompt.

\begin{figure}[t]
    \centering
    \includegraphics[width=0.85\linewidth]{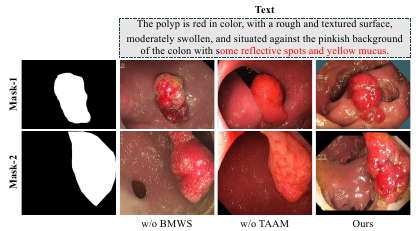}
    \vskip -15pt
    \caption{
    Visualization of the ablation experiments.
    }
    \label{fig:figure4}
    \vskip -15pt
\end{figure}

To conduct a more comprehensive evaluation, we implemented a user study questionnaire. We selected 30 groups of images for comparison and invited 15 clinical experts from the First Affiliated Hospital of Sun Yat-sen University to participate in the survey. 
Participants were required to rate the synthesized images based on three criteria: image fidelity, mask accuracy, and text accuracy. The average scores for each criterion are presented in Table \ref{table2}. Our method outperforms the other methods in all aspects.

To further verify the effectiveness of our synthetic images, we evaluated our method against state-of-the-art methods for the polyp segmentation task.
We generated the same number of samples as the diffusion training set using the original masks, which we then combined to create a new downstream training dataset. 
The experimental results presented in Table \ref{table3} highlight the effectiveness of our method in training improved downstream models that achieve superior performance.

We also conduct ablation visualization experiments to demonstrate the effectiveness of the Blur Mask Weighting Strategy (BMWS) and the Text-Aware Attention Mechanism (TAAM). The results are illustrated in Fig. \ref{fig:figure4}. The combination of BMWS and TAAM significantly enhances the quality of the synthesized images and text alignment.

\section{Conclusion}
\label{sec:Conclusion}
In this paper, we present CCIS-DIFF, a generative model that leverages a Stable Diffusion prior for the controlled colonoscopy image synthesis. 
We begin by developing a blur mask weighting strategy to seamlessly integrate the generated polyp with the colonic mucosa, along with a text-aware attention mechanism to address the issue of neglecting text prompt.   
Additionally, we introduce a multi-modal colonoscopy dataset created by large language models.
Extensive experiments across various settings demonstrate the superior performance of CCIS-DIFF.

\section{Compliance with Ethical Standards}
This research study was conducted retrospectively using human subject data made available in open access. Ethical approval was not required as confirmed by the license attached with the open access data.


\bibliographystyle{IEEEbib}

\end{document}